\documentclass[runningheads]{llncs}
\usepackage{graphicx}
\usepackage{xcolor}

\begin{document}
\title{Quick and Easy Time Series Generation with Established Image-based GANs}
\titlerunning{Time series Generation with GANs}
%
\author{Eoin Brophy\orcidID{0000-0002-6486-5746} \and
Zhengwei Wang\orcidID{0000-0001-7706-553X} \and
Tom\'as E. Ward\orcidID{0000-0002-6173-6607}} 
\authorrunning{Brophy et al.}
%
\institute{Insight Centre for Data Analytics, \newline
Dublin City University, Glasnevin, Dublin, Ireland}
\maketitle              
%

\begin{abstract}
In the recent years Generative Adversarial Networks (GANs) have demonstrated significant progress in generating authentic looking data. In this work we introduce our simple method to exploit the advancements in well established image-based GANs to synthesise single channel time series data. We implement Wasserstein GANs (WGANs) with gradient penalty due to their stability in training to synthesise three different types of data; sinusoidal data, photoplethysmograph (PPG) data and electrocardiograph (ECG) data. The length of the returned time series data is limited only by the image resolution, we use an image size of 64x64 pixels which yields 4096 data points. We present both visual and quantitative evidence that our novel method can successfully generate time series data using image-based GANs. 

\end{abstract}
\section{Introduction}
Time series data is abundant in many fields of study from health and medical recordings to financial and weather statistics. However, the issue with time series data, particularly in terms of physiological data is that of privacy. Following the introduction of General Data Protection Regulation (GDPR) in Europe in 2018, the access to data has become a hindrance in the progression of scientific studies, notably in fields relating to medical and physiological data. 

However, open source data sets are readily available and allow for comparisons of machine learning models and dissemination of results. Often, these data sets may not satisfy criteria for a study and even occasionally missing some data. Using private data sets results in experiments that are difficult to replicate and this leads to a lack of development in scientific progress. Recent research has tried to synthesise time series data using both synthesisers and machine learning models to some degree of success. 

\begin{figure}
\centering
\includegraphics[width=1\textwidth]{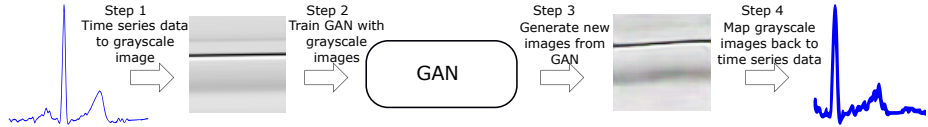}
\caption{Pipeline of GAN model. Step one involves segmenting the raw time series data into suitable windows of fixed length, and sampling the amplitude of the signal to map on to an RGB grayscale value. The RGB grayscale values are rasterised into an image. Step two involves training the GAN using the rasterised images. Step three uses the trained GAN is used to synthesise new images. Step four involves transforming the images back to their corresponding time series array.} \label{fig1}
\end{figure}

In this paper we looked to synthesise three different types of time series data; sinusoidal, photoplethysmograph (PPG) and electrocardiograph (ECG). Our method begins with segmenting the data into suitable windows of fixed length. We then sample the amplitude of the signal and map the amplitude to an RGB grayscale value. The array of RGB values are rasterised into an image that can be used to train a Generative Adversarial Network (GAN). The trained GAN is then capable of synthesising new rasterised images which can then be transformed back to time series using our transform, see Fig.\ref{fig1}. The term `rasterised image' used throughout this paper is referring to the type of images shown in Fig.\ref{fig2}.

The contributions of our approach allows for easily deployable, single channel time series synthesis using generative adversarial networks. The method used in this paper is not claiming to be the solution to time series synthesis/imputation. However, it is a novel and easily implemented solution to a difficult task. While the current recurrent GANs have shown success in the area of time series synthesis, they are in a relatively early phase and do not yet have optimal design stability due to difficulties in the training phase. We suggest this method as an early stage approach in generating time series using image-based GANs; a `for beginners' solution.

\section{Related Work}
Few studies have used GANs to produce time series data as they have mainly been developed for the generation of images. Some recent results showed promise in synthesising time series data \cite{Esteban2017,Hartmann2018}. Hartmann et al. \cite{Hartmann2018} used channel FCC4h recorded from a 128-electrode electroencephalograph (EEG) system down sampled to 250 Hz as training samples for their EEG-GAN framework. With this they demonstrated the ability of their EEG-GAN for the generation of time series EEG data up to 768 time samples.

An important advance was introduced by Esteban et al. \cite{Esteban2017}, a method of synthesising time series using recurrent conditional generative adversarial networks (RCGAN). They synthesised both time series sinusoidal data and physiological data; oxygen saturation, heart rate, respiratory rate and mean arterial pressure. The data was gathered from the eICU Collaborative Research Database. The authors state the length of their generated data sequences as 30 data points.

Our work seeks to simplify the current methods of generating time series data using mature image-based GAN frameworks which have shown stability in training. The simplicity of our approach allows for highly flexible lengths of data, limited only by the image resolution.

\section{Methodology}
\subsection{Data Collection} 
The data used in this experiment was gathered from PhysioNet \cite{PhysioNet} and generated using a sinusoidal generator. The ECG data was gathered from an experiment conducted by Garc\'ia-Gonz\'alez et al. \cite{Garcia-Gonzalez2013} and the PPG from Pimentel et al. \cite{Pimentel2017}. The sinusoidal data was generated in Python using NumPy's built-in sine function.

\subsection{Data Preparation}
\begin{figure}
    \centering
    \includegraphics[width=1\textwidth]{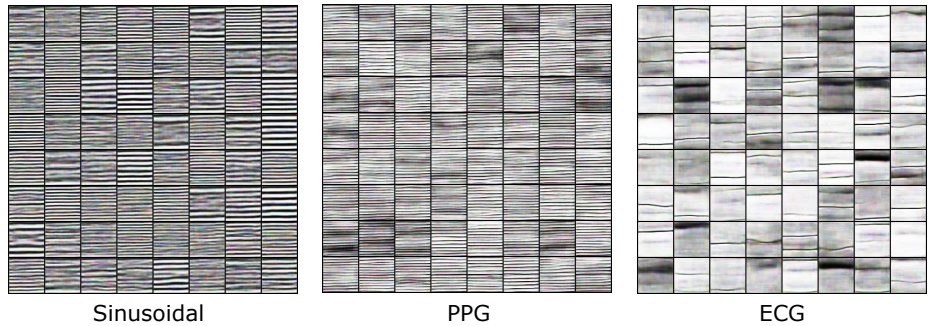}
    \caption{A square grid of 8x8 (64) examples of rasterised RGB images. These images are taken from the training set. The amplitude and frequency information preserved in these images allow for successful reconstruction to time series data.}
    \label{fig2}
\end{figure}

The number of training images used for each signal type was; Sinusoidal-GAN 2500 images, PPG-GAN 2320 images and ECG-GAN 4880 images. The sinusoidal data was generated using signals of varying amplitude and frequency. The PPG was gathered using a sampling frequency of 125Hz, we then up-sampled it to 256Hz and chose data windows of 16 seconds as this would give us a 4096 (256x16) vector that could be rasterised easily into a 64x64 RGB image. The ECG data was sampled at 5kHz and we took one second windows of the time series data (which allowed us to capture at least one QRS complex per window) that was to be rasterised into it’s RGB representation. See Fig.\ref{fig2} for examples of our rasterised sinusoidal, PPG and ECG data. 

\subsection{The Network}
Due to the difficult nature of training GANs we decided to use a Wasserstein Generative Adversarial Network (WGAN) with gradient penalty. The model architecture used in this experiment was the same as the architecture used in \cite{Gulrajani2017}. A WGAN requires that the critic lies within the space of 1-Lipschitz function which is enforced through a weight clipping in a standard WGAN \cite{Arjovsky2017}. The weight clipping can result in vanishing or exploding gradients if the clipping threshold is not tuned precisely. The gradient penalty is another way to enforce the weight clipping and bypasses the training issues that other GANs suffer from \cite{Gulrajani2017}. Each WGAN was trained for 300 epochs, the results of training the WGANs with gradient penalty can be found in the sections to follow.

\section{Results}
\begin{figure}
\centering
\includegraphics[width=1\textwidth]{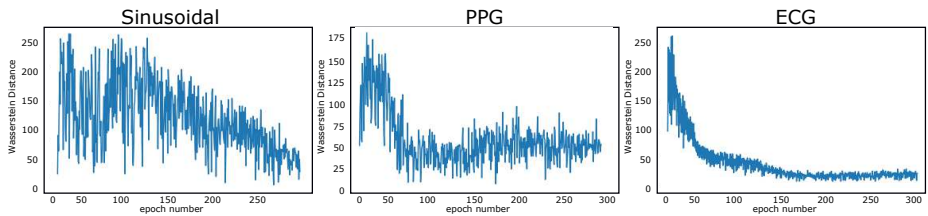}
\caption{Wasserstein-1 distances for each signal} \label{fig4}
\end{figure}

 Shown in table \ref{table:1} below is the Frech\'{e}t Inception Distance (FID) and maximum-mean discrepancy (MMD) for images generated at epoch 300. The WGAN was not optimised using these distance metrics but they give an indication that the distance between the source and target probability distributions is quite large. However, when optimised with the Wasserstein-1 distance, we can observe the distance reducing on each iteration of training, see Fig.\ref{fig4}. This demonstrates that the generated and real probability distribution distances are getting closer to one another.
 
 The features used in calculating the FID, MMD and Wasserstein distances were the raw pixel values of the images. These metrics allow us to quantify the GAN performance and demonstrate the feasibility of this study.

\begin{table}[h!]
\centering
\begin{tabular}{c | c | c} 
   & FID & MMD \\
 \hline
 Sinusoidal & 72.53 & 71.75 \\ \hline
 PPG & 86.74 & 86.55 \\ \hline
 ECG & 109.17 & 111.36 \\
\end{tabular}
\caption{FID and MMD for the trained GAN}
\label{table:1}
\end{table}

\begin{figure}
\includegraphics[width=1\textwidth]{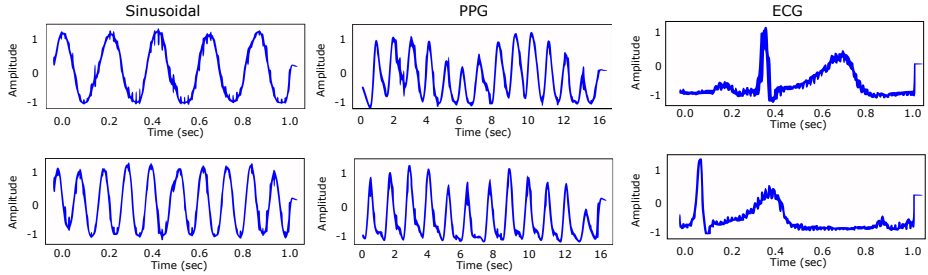}
\caption{Synthesised Sinusiodal, PPG and ECG data. The bottom plots are filtered using a simple low pass filter with different cut-off frequencies. Further plots can be viewed in the Appendix.} \label{fig5}
\end{figure}

In Fig.\ref{fig5} two examples of each synthesised signal from our WGANs can be seen. These time series signals are returned from our RGB to time series transform and exhibit the characteristics of each signal it was trained on with some additional noise artefacts.

\section{Discussion}
\begin{figure}
\includegraphics[width=1\textwidth]{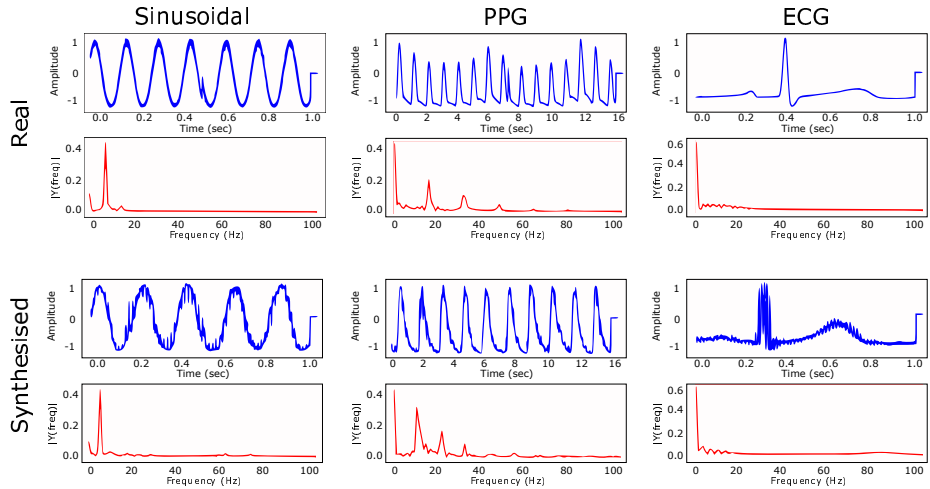}
\caption{FFT of each signal type, both real and synthesised. $ \vert Y(freq) \vert $ denotes the magnitude of the frequency.} \label{fig6}
\end{figure}

The network successfully generated rasterised images that look similar to the training data set. Using the rasterised image to time series conversion function shows the generated time series suffers from some high frequency noise. Applying a low-pass filter will remove this noise and return a clean and fully synthesised time series sequence. Fig.\ref{fig6} shows the fast fourier transform (FFT) for each signal, both real and synthesised. The frequency domain demonstrates the similarity between these signals for each of the data sets. 

\subsection{Limitiations}
Of course, the generated time series sequences, especially the ECG signal may be too short for most applications. Therefore, we suggest concatenating generated sequences or selecting a reduced sample frequency when collecting the data. 

We do not claim that this method should replace the current standards in time series generation but rather it should be viewed as a novel and simple approach for single channel time series generation.

\section{Conclusion}
We have designed a procedure that is capable of exploiting image-based GANs for generating time series data of flexible length. This can be applied to any single channel time series data and with some future experimentation, potentially multichannel data. We propose that our approach represents a useful way of synthesising time series data, circumventing the problems inherent with private data and facilitating the development large synthetic databases. The synthetic nature of the data will allow for the generation of and publication on large synthetic databases more readily, further advancing the scientific progress.

\subsubsection{Acknowledgements} This work is part-funded by Science Foundation Ireland under grant numbers 17/RC-PhD/3482 and SFI/12/RC/2289. We gratefully acknowledge the support of NVIDIA Corporation with the donation of the Titan Xp used for this research. The authors would also like to thank Darragh Walsh for proofreading this paper. 

%
%

%
%
%
%

\bibliography{references} 
\bibliographystyle{splncs04}

\section*{Appendix}

\begin{figure}
\includegraphics[width=1\textwidth]{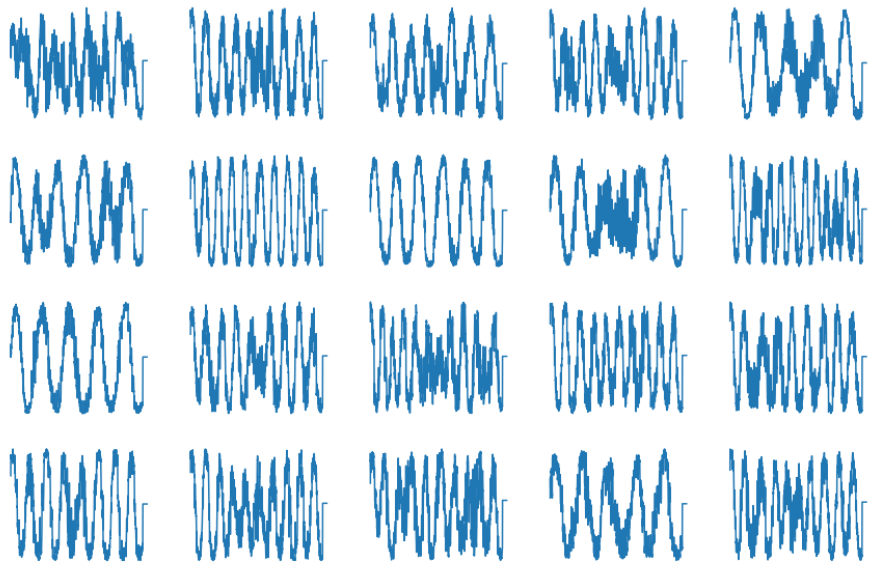}
\caption{Further synthesised time series plots of sinusoidal data} \label{fig7}
\end{figure}

\begin{figure}
\includegraphics[width=1\textwidth]{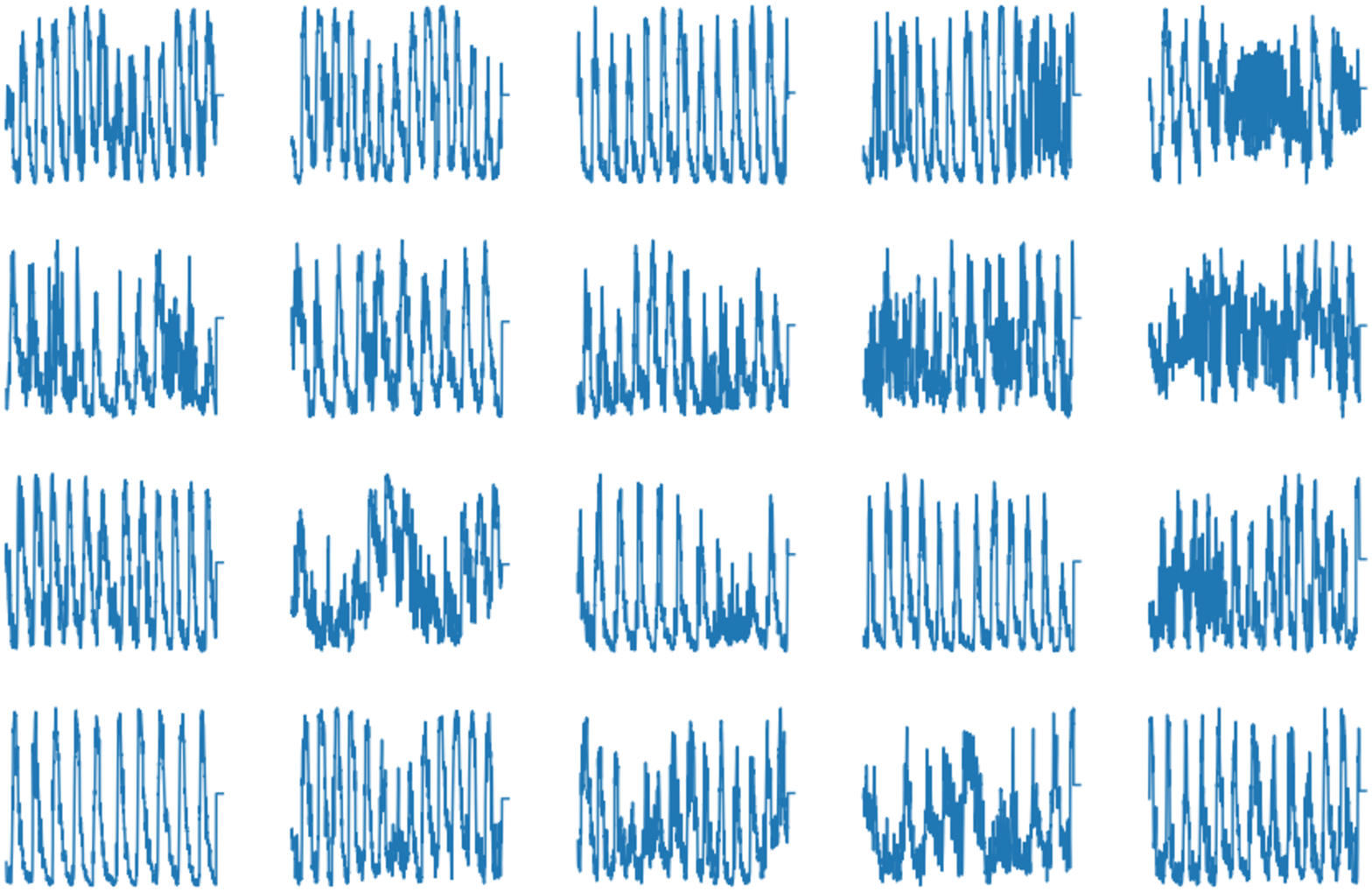}
\caption{Further synthesised time series plots of PPG data} \label{fig8}
\end{figure}

\begin{figure}
\includegraphics[width=1\textwidth]{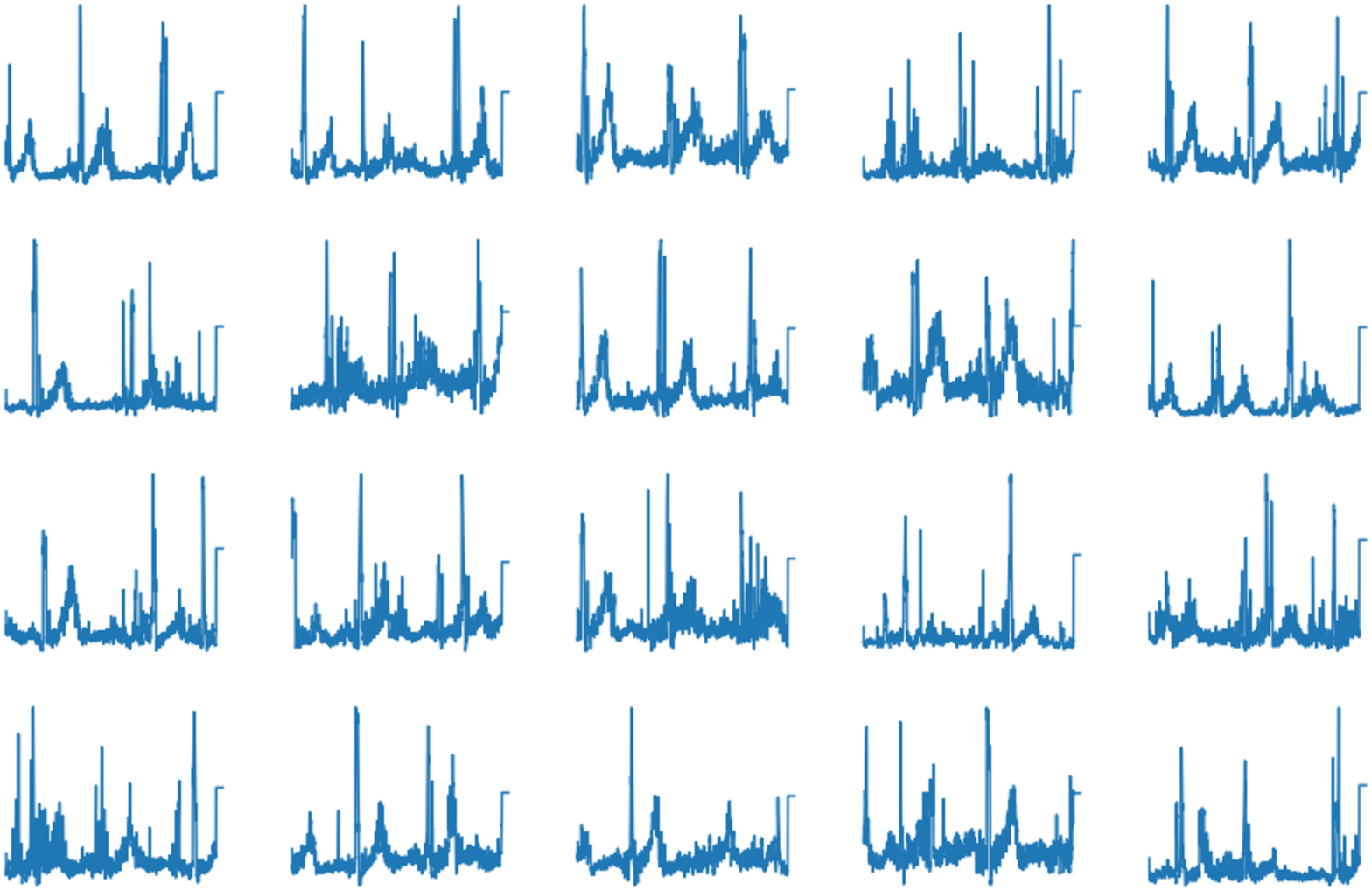}
\caption{Further synthesised time series plots of ECG data} \label{fig9}
\end{figure}

\end{document}